# Research on Event Accumulator Settings for Event-Based SLAM

KUN, XIAO

Beijing Institute of Aerospace Systems Engineering

GUOHUI, WANG

China Academy of Launch Vehicle Technology

YI, CHEN

Beijing Institute of Aerospace Systems Engineering

YONGFENG, XIE

Beijing Institute of Aerospace Systems Engineering

HONG, LI

Beijing Institute of Aerospace Systems Engineering

SEN, LI

China Academy of Launch Vehicle Technology

Event cameras are a new type of sensors that are different from traditional cameras. Each pixel is triggered asynchronously by event. The trigger event is the change of the brightness irradiated on the pixel. If the increment or decrement of brightness is higher than a certain threshold, an event is output. Compared with traditional cameras, event cameras have the advantages of high dynamic range and no motion blur. Accumulating events to frames and using traditional SLAM algorithm is a direct and efficient way for event-based SLAM. Different event accumulator settings, such as slice method of event stream, processing method for no motion, using polarity or not, decay function and event contribution, can cause quite different accumulating results. We conducted the research on how to accumulate event frames to achieve a better event-based SLAM performance. For experiment verification, accumulated event frames are fed to the traditional SLAM system to construct an event-based SLAM system. Our strategy of setting event accumulator has been evaluated on the public dataset. The experiment results show that our method can achieve better performance in most sequences compared with the state-of-the-art event frame based SLAM algorithm. In addition, the proposed approach has been tested on a quadrotor UAV to show the potential of applications in real scenario. Code and results are open sourced to benefit the research community of event cameras[1].

---

[1] https://github.com/robin-shaun/event-slam-accumulator-settings or https://gitee.com/robin_shaun/event-slam-accumulator-settings

## 1 INTRODUCTION

For mobile robots, it is important to understand the surrounding environment and estimate their poses [1]. While being cost-effective, visual Simultaneous Localization and Mapping (SLAM) and Visual Odometry (VO) have been shown to achieve high accuracy and robustness in many applications [2–4]. By adding an Inertial Measurement Unit (IMU), the accuracy and robustness of VO can be further improved, which is called Visual Inertial Odometry (VIO) [5–7]. However, there are still scenarios which are challenging for visual-inertial systems such as under very fast motions or in scenes with High Dynamic Range (HDR) of illumination.

Event cameras are new sensor types that have a huge potential in addressing aforementioned limitations due to their extremely high temporal resolution and their HDR operative range [8, 9]. In contrast to traditional frame-based cameras, which periodically output intensity values for all pixels, these sensors output event tuples $e = \{t, x, y, p\}$, where $t$ is the timestamp of the event, $x$, $y$ are the image coordinates, and $p$ is the event polarity. Such an event is triggered only in case a pixel's intensity change is larger than a threshold, with $p$ being the direction of that change. Consequently, in a static scene, events are only generated when the camera is in motion. The events form an asynchronous stream of data that provides reliable information even in the presence of fast motion.

To apply or adapt conventional frame-based SLAM algorithms, a direct and efficient way is to accumulate events to frames. EVO [10], ESVO[2] [11] and EFVIO[12] are three representative works, which solve monocular, stereo and visual-inertial SLAM problem respectively and all of them are real-time and high-accuracy. Although they are both event frame based SLAM, the way to accumulate events are different. Indeed, there are many methods and settings to accumulate events to frames, but as we know, no paper researches how to adjust the accumulator settings for event-based SLAM in detail.

The remainder of this paper is organized as follows: In Section 2, the event accumulator and the event frame are introduced. In Section 3, the event generation model is presented as it is the theory basis of the analysis of the event accumulator settings. In Section 4, different event accumulating settings, including slice method, processing method for no motion, polarity usage, decay function and event contribution, are analyzed. In Section 5, accumulated event frames are fed to the traditional SLAM system to construct an event-based SLAM system and the strategy of accumulator settings are evaluated on both the public dataset and the UAV flight experiments. In Section 6, conclusion is made and future work is discussed.

## 2 EVENT ACCUMULATOR AND EVENT FRAME

Events flow along the time axis, forming the temporal and spatial distribution of events, as shown in Figure 1. This representation is hard to be used by the conventional computer vision algorithm. Therefore, events are often processed and transformed into alternative representations that facilitate the extraction of meaningful information ("features") to solve a given task.

The events in a spatio-temporal neighborhood can be converted accumulating polarity pixel-wise into an image (2D grid) that can be fed to image-based computer vision algorithms [14]. This processing unit is called event accumulator and this kind of image is called event frame. Event frames are widely used in the event-based vision because (i) they are a simple way to convert an unfamiliar event stream into a familiar 2D representation containing spatial information about scene edges, which are the most informative regions in natural images, (ii) they inform not only about the presence of events but also about their absence (which is informative), (iii) they have an intuitive interpretation (e.g., an edge map, a brightness increment image).

---

[2] ESVO exploits time surface, which encodes spatio-temporal information to a frame. Time surface is also a kind of event frame in nature. Jiao et al. compares the time surface and the traditional event frame for event-based SLAM [13] and our paper focus on the research of the traditional event frame.

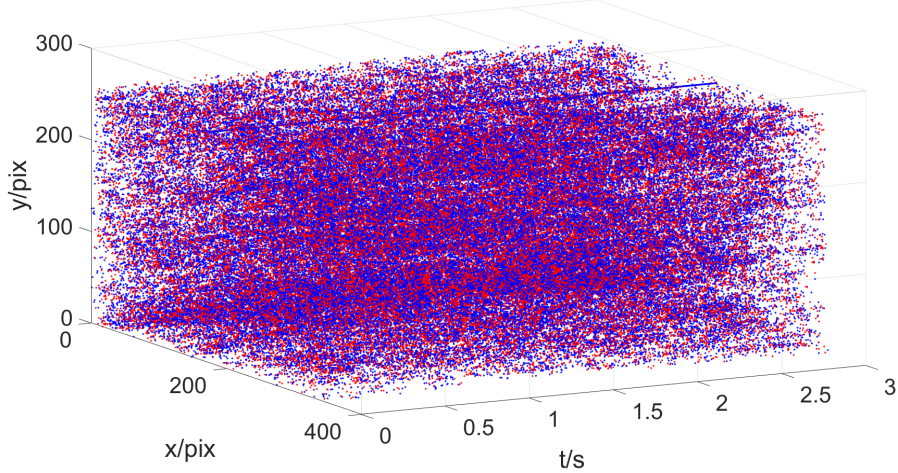

Figure 1: Events in space time, colored according to polarity (positive in red, negative in blue)

## 3 EVENT GENERATION MODEL

An event camera has independent pixels that respond to changes in their log photocurrent $L \doteq \log(I)$ ("brightness"). Specifically, in a noise-free scenario, an event $e_k \doteq (\mathbf{x}_k, t_k, p_k)$ is triggered at pixel $\mathbf{x}_k \doteq (x_k, y_k)^\top$ and at time $t_k$ as soon as the brightness increment since the last event at the pixel, i.e.

$$\Delta L(\mathbf{x}_k, t_k) \doteq L(\mathbf{x}_k, t_k) - L(\mathbf{x}_k, t_k - \Delta t_k) \tag{1}$$

reaches a temporal contrast threshold $\pm C$, i.e.,

$$\Delta L(\mathbf{x}_k, t_k) = p_k C \tag{2}$$

where $C > 0$, $\Delta t_k$ is the time elapsed since the last event at the same pixel, and the polarity $p_k \in \{+1, -1\}$ is the sign of the brightness change [8].

Assuming constant illumination, linearizing (1) and using the brightness constancy assumption one can show that events are caused by moving edges. For small $\Delta t$, the intensity increment can be approximated by:

$$\Delta L \approx -\nabla L \cdot \mathbf{v} \Delta t = -\nabla L \cdot \Delta \mathbf{x} \tag{3}$$

that is, it is caused by a brightness gradient $\nabla L(\mathbf{x}_k, t_k) = (\partial_x L, \partial_y L)^\top$ moving with velocity $\mathbf{v}(\mathbf{x}_k, t_k)$ on the image plane, over a displacement $\Delta \mathbf{x} \doteq \mathbf{v} \Delta t$. Eq.(3) is quite important for SLAM because it clarifies how the scene and ego-motion influence event output ($\nabla L$ and $\Delta \mathbf{x}$ respectively). The next section will discuss Eq.(3) from different perspectives.

## 4 EVENT ACCUMULATOR SETTINGS

The accumulator module takes events and prints them onto a frame. Removal of the information is done with a configurable decay function. This section will introduce different accumulator settings and the principle to adjust the settings.

### 4.1 Slice Method

A spatio-temporal window of events is called a slice. There are two basic methods to create a slice, slice by number or slice by time, but both of them are not suitable for SLAM systems.

The first method is to create slices according to a fixed number of events. For example, the fixed number is 5000. When the number of new events reaches 5000, a new slice is created and a new event frame is output. Because events are output at variable intervals, this method output event frames aperiodically, which is incompatible with conventional stereo SLAM systems. Stereo SLAM usually strictly limits the time difference of frames from different cameras, and event frames created by this method are likely not to meet this time difference requirement. In addition, the event frames created by this method also face the problem of time stamp synchronization when fused with data from other sensors, such as IMUs and lidars.

Unlike slice by number, the second method is to create slices and outputs frames at regular intervals. However, it is also not suitable for SLAM systems. Assuming that the scene changes little, more fast motion will cause more events at a same interval. Figure 2 shows three event frames with the event camera moving at different speeds. These three event frames come from a similar scene, but look quite different. It is quite hard to associate these three frames and therefore SLAM systems cannot work well.

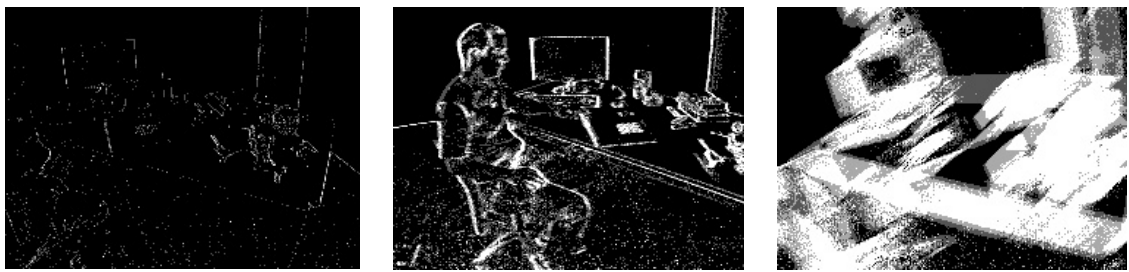

(a) Event camera moving slowly  (b) Event camera moving at a normal speed  (c) Event camera moving fast

Figure 2: Three event frames from a similar scene using slice by time, but looking quite different

Ultimate SLAM [15] combines these two methods and we call this method slice by time and number. Upon the time $t_k$ when an event frame will be published, a new slice $S_k$ is created (Figure 3). The $k^{th}$ slice is defined as the set of events $S_k = \{e_{j(t_k)-N+1}, \ldots, e_{j(t_k)}\}$, where $j(t_k)$ is the index of the first event whose timestamp $t_j < t_k$, and $N$ is the window size parameter. Note that the duration of each slice is inversely proportional to the event rate.

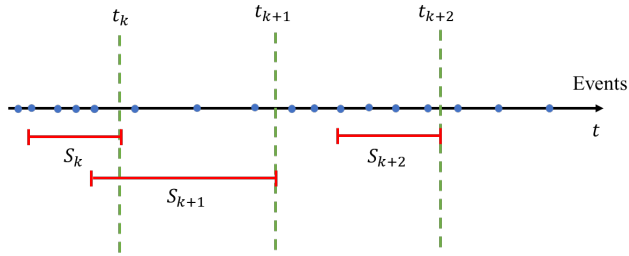
Figure 3: Slice by time and number (Blue dots corresponding to events, the dashed green lines corresponding to the times when event frames are published, and the bounds of the slices marked in red)

Slice by time and number can output event frames at regular intervals, which is compatible with conventional SLAM systems. Unlike slice by time, slices create by this method maintain the same window size. According to Eq.(3), assuming that the scene (described by $\nabla L$) changes little, the same number of events are caused by similar displacement. This means that whether moving fast or not, the output frames look similar, which are quite important for feature tracking or matching. Figure 4 shows three event frames with the event camera moving at different speeds, but looking similar. Therefore, slice by time and number is recommended for event frame based SLAM.

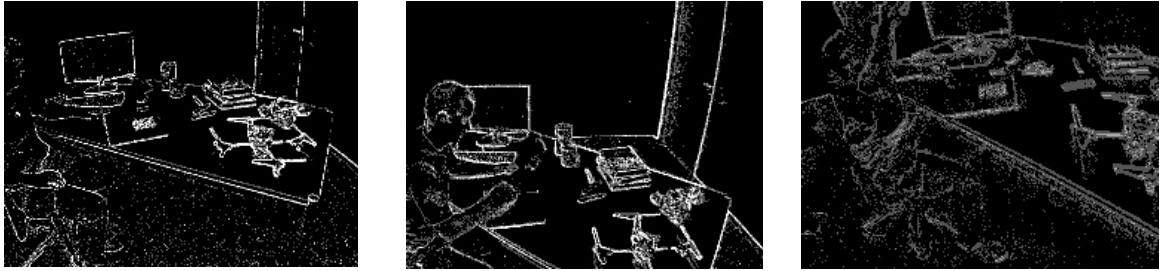

(a) Event camera moving slowly  (b) Event camera moving at a normal speed  (c) Event camera moving fast

Figure 4: Three event frames from a similar scene using slice by time and number, looking similar

For the method of slice by time and number, there are two parameters, time interval and event window size. The time interval decides the event frame output frequency, so it depends on the processing speed of the SLAM system. The event window size depends on the camera resolution and the scene structure a lot but depends on the camera motion speed little. More texture the scene structure has, more complex it is. A factor, event number per pixel, is used to represent the complexity of scene structure.

$$\text{Event window size} = \text{event number per pixel} \times \text{frame width} \times \text{frame height}$$

Event window size is a key parameter for event frame based SLAM. It should be adjusted according to the event frame. The criterion is that the edges should be sharp (without motion blur) and the event frame should fully reflect the scene structure. Figure 5 shows that when the event window size is too large, edges are not sharp and have motion blur, and when the event window size is too small, event frame cannot fully reflect the scene structure.

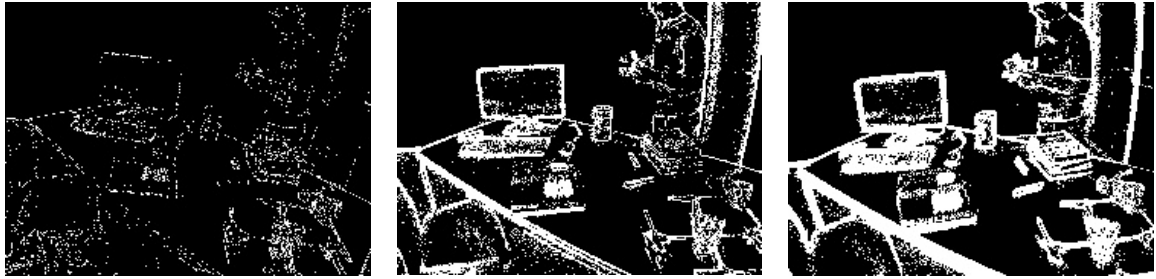

(a) Event window size 2000 (small)     (b) Event window size 10000 (proper)     (c) Event window size 30000 (large)

Figure 5: Event frame comparison with different event window sizes

### 4.2 Processing Method for No Motion

According to the principle of event cameras, when the camera doesn't move, there is no event output in theory. However, there are still noise events in practice. Using slice by time and number, more and more pixels are filled with noise when event camera keeps still, shown as Figure 6.

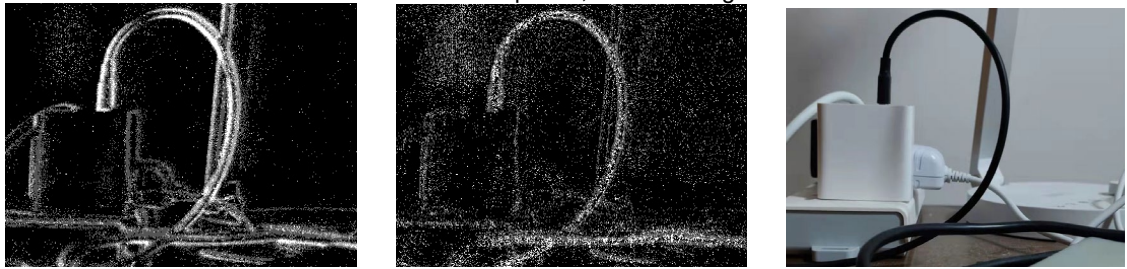

(a) Camera moving     (b) Keeping still for 0.5s     (c) Conventional color frame

Figure 6: Event frame becomes noisy when camera keeps still

To tackle this problem, a simple but effective method is proposed. A threshold is set, and when the number of events at the interval (i.e., $t_k - t_{k-1}$) is less than the threshold, the old event frame will copy to the new event frame, which means that the camera keeps still at this interval. This method is proved to be effective in the experiment.

### 4.3 Polarity Usage

According to Section 3, the polarity $p_k \in \{+1, -1\}$ is the sign of the brightness change. For image reconstruction [16], the polarity is useful. However, for SLAM, polarity can cause trouble. When the event camera moves in another direction suddenly, the polarity will be changed, which will make the pixel values of edges in the event frame change suddenly, shown as Figure 7. This will influence the image matching seriously, which can make SLAM systems fail. Therefore, the event accumulator should rectify polarities (usually both to 1) first and then accumulate events to event frames.

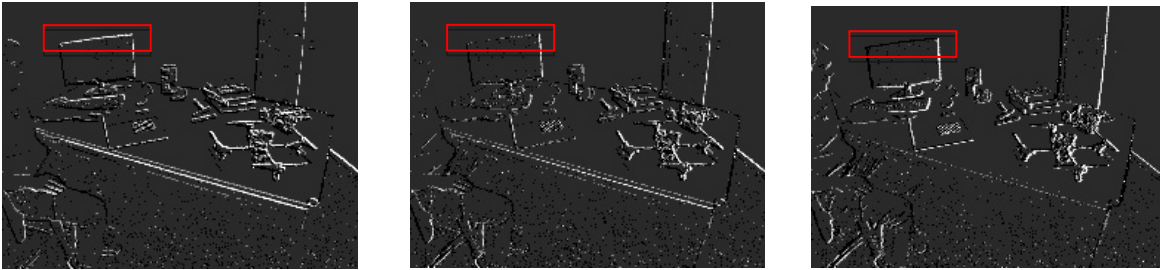

(a) Camera moves in one direction   (b) Camera starts moving in another direction   (c) Camera moves in another direction

Figure 7: Pixel values of edges in the event frame change, and the edge in the red box changes the most obviously[3]

### 4.4 Decay Function

Three slice methods introduced above all use one event window, which is one kind of event accumulator. Generally, an event accumulator can use all event windows to reconstruct an absolute brightness image. Events represent brightness changes, and so, in ideal conditions (noise-free scenario, perfect sensor response, etc.), integration of the events yields "absolute" brightness. However, real event cameras are noisy and differ significantly from the ideal camera model, which renders the reconstruction problem ill-posed. Naive integration of the event stream leads to very fast degradation of image quality due to accumulating noise [16]. Although linear or exponential decay can suppress noise effectively, shown as Figure 8, the reconstruction effect is not so good[4]. According to our experiment result, this kind of reconstruction is not suitable for SLAM systems. In fact, three slice methods introduced above use a step decay function. The step decay function only reserves edges in the event frame, which is enough for SLAM systems.

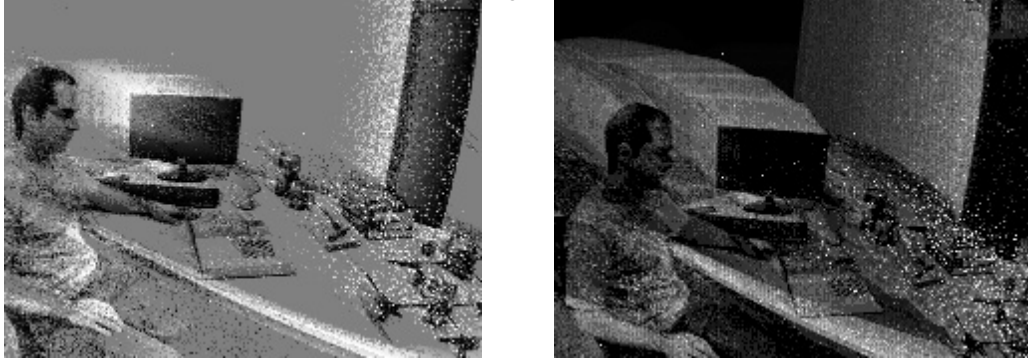

(a) Linear decay                (b) Exponential decay

Figure 8: Event frames from accumulator with linear and exponential decay functions

---

[3] Notice that the background of event frames in Figure 7 is not black but gray, which is different from Figure 2, 4, 5 and 6. Generally, for the event accumulator without rectifying polarities, the initial pixel value is the half of the max pixel value, rather than zero.

[4] Image reconstruction is an important research area for event-based vision, some researches can realize real-time and high-quality reconstruction, such as [17, 18]. High-quality reconstructed brightness images are suitable for SLAM systems, but it's not the topic in this paper.

## 4.5 Event Contribution

Event contribution is the contribution an event has onto the image. If an event arrives at a position $(x, y)$, the pixel value in the frame at $(x, y)$ gets increased / decreased by the value of event contribution, based on the events polarity. For simplification, normalization is used, which means that event contribution ranges from 0 to 1. When event contribution is 0.5, two events with positive polarity make the pixel value 255.

We tested different event contribution settings on a special picture, shown as Figure 9. Figure 9 (a) has two edges with different sharpness. If event contribution is 1, the event frame is a binary image, and cannot distinguish these two edges, shown as Figure 9 (b). In theory, the less the event contribution is, the larger the bit depth is, which is better for feature tracking and matching. However, the event count in a same pixel is small generally. Therefore. too little event contribution makes edges weak, like Figure 9 (h). Because different event cameras and different scenes make event count statistical contribution different., the event contribution settings should be adjusted according to the image quality.

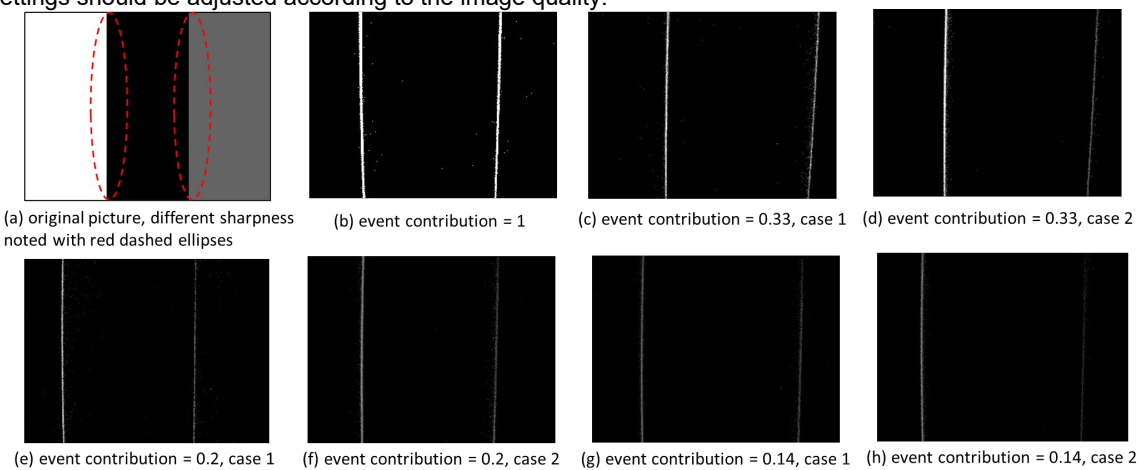

(a) original picture, different sharpness noted with red dashed ellipses
(b) event contribution = 1
(c) event contribution = 0.33, case 1
(d) event contribution = 0.33, case 2
(e) event contribution = 0.2, case 1
(f) event contribution = 0.2, case 2
(g) event contribution = 0.14, case 1
(h) event contribution = 0.14, case 2

Figure 9: Event frames with different event contribution settings

## 5 EXPERIMENT AND EVALUATION

For experiment verification, accumulated event frames are fed to the traditional SLAM system to construct an event-based SLAM system. To verify that the proposed method meets general need, we choose a special SLAM system, a VIO system, which involves the synchronization of time stamps between the image and the IMU. VINS-Mono [5] is chosen for its stabilized performance and good open-source community[5]. The experiment consists of two parts, dataset test and UAV flight experiments. The former is to compare the proposed method with the state-of-the-art event-based VIO system and the latter is to test its performance on an onboard computer.

---

[5] https://github.com/HKUST-Aerial-Robotics/VINS-Mono

## 5.1 Dataset Test

We evaluate the proposed method quantitatively on the Event Camera Dataset [19], which features various scenes with ground truth tracking information. In particular, it contains extremely fast motions and scenes with very high dynamic range, recorded with the DAVIS240 [9] sensor. We exclude the rotational only datasets, as well as the datasets without inertial measurements.

The DAVIS240 sensor embeds a $240 \times 180$ pixels event camera with a 1kHz IMU. Events and IMU measurements are synchronized on hardware. The IMU is delayed by a constant time offset in the order of 2.5ms compared to the events (because of the low-pass filter of the IMU). We estimated this delay using Kalibr [20].

To evaluate the results, the estimated and ground truth trajectories are aligned with a 6-DOF transformation in SE3, using 5 seconds of the trajectory (starting at second 3 and ending at second 8). Then, we compute the mean position error (Euclidean distance) and the yaw error. We use rpg_trajectory_evaluation toolbox [21] to realize algorithm evaluation. Table 1 shows the comparison between our approach and the state-of-the-art event frame based VIO system [12]. Table 2 shows the event accumulator settings in VINS-Mono we used for the 10 sequences. The frequency of VINS-Mono is set to the same as the event frame frequency, 30Hz. Because all the sequences do not include no motion case, the no motion threshold is not shown in Table 2. Our approach is better than the current state-of-the-art algorithm on 8 sequences. Performance on shape_6dof and shape_translation is similar to the current state-of-the-art.

Table 1: Comparison between our approach and the state-of-the-art [12], bold numbers represent better performance

| Sequence | Proposed | | State-of-the-art | |
|---|---|---|---|---|
| | Mean Position Error (%) | Mean Yaw Error (deg/m) | Mean Position Error (%) | Mean Yaw Error (deg/m) |
| dynamic_6dof | **0.29** | **0.02** | 0.56 | 0.41 |
| dynamic_translation | **0.24** | **0.05** | 0.39 | 0.06 |
| shape_6dof | 0.55 | **0.08** | **0.42** | 0.18 |
| shape_translation | **0.45** | 0.11 | 0.46 | **0.10** |
| boxes_6dof | **0.29** | **0.05** | 0.36 | 0.11 |
| boxes_translation | **0.28** | **0.03** | 0.31 | 0.08 |
| poster_6dof | **0.21** | **0.02** | 0.40 | 0.16 |
| poster_translation | **0.13** | **0.04** | 0.46 | 0.10 |
| hdr_poster | **0.30** | **0.05** | 0.33 | 0.19 |
| hdr_boxes | **0.27** | **0.02** | 0.59 | 0.20 |

Table 2: Event accumulator settings in VINS-Mono for the dataset test

| Sequence | Slice method | Polarity usage | Decay function | Event contribution | Event window size |
|---|---|---|---|---|---|
| dynamic_6dof | By time and number | No | Step | 0.2 | 10000 |
| dynamic_translation | | | | 0.33 | 10000 |
| shape_6dof | | | | 0.2 | 3000 |
| shape_translation | | | | 0.2 | 3000 |
| boxes_6dof | | | | 0.2 | 15000 |
| boxes_translation | | | | 0.2 | 15000 |
| poster_6dof | | | | 0.2 | 10000 |
| poster_translation | | | | 0.33 | 10000 |
| hdr_poster | | | | 0.2 | 10000 |
| hdr_boxes | | | | 0.2 | 20000 |

### 5.2 UAV Flight Experiments

In order to show the potential of our proposed method in a real scenario, we did UAV flight experiments. A quadrotor UAV with Pixhawk controller [22] and NVIDIA Jetson Xavier NX onboard computer is used, shown as Figure 10(a). Pixhawk controller armed with PX4 Flight Stack[6] is used for the UAV control. The proposed event accumulator and VINS-Mono are all computed on Xavier NX which has 6-core NVIDIA Carmel ARMv8.2 64-bit CPU and 384 NVIDIA CUDA cores under Ubuntu 18.04 and ROS [23] system. A DAVIS346 sensor, which embeds a 346×260 pixels event camera with a 1kHz IMU, mounted on the middle of the quadrotor UAV, looking forwards. A motion capture system (mocap) is used to obtain the pose groundtruth. Because the active IR emitters on the mocap cameras can influence DAVIS346 greatly, they are closed and retroreflective markers are not used. Instead, some IR LEDs are mounted on the top of the quadrotor UAV to supply IR for the mocap.

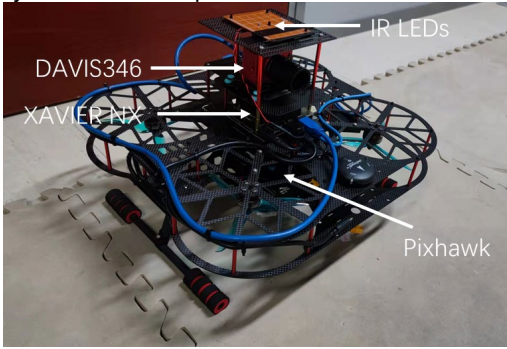 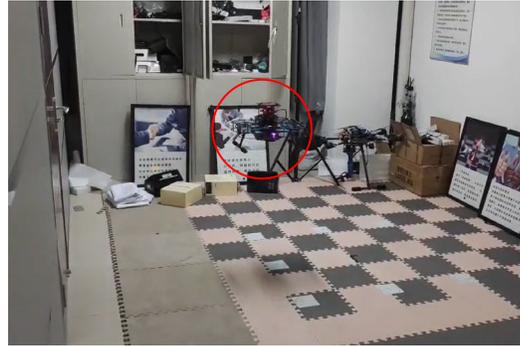

(a) Quadrotor UAV experiment platform      (b) Experiment scenario (UAV marked in the red circle)

Figure 10: Quadrotor UAV experiment platform and experiment scenario

In this experiment, the event contribution is set to 0.5 instead of 0.2 for the event generating efficiency not good as the Event Camera Dataset. The event window size is set to 20000 and the no motion threshold is 200. Other settings are the same as the settings in Table 2. According to statistics, under this setting, it takes an average of about 7ms to generate an event frame on Xavier NX. Almost half of the time consumption comes from grabbing event from DAVIS, which can be optimized by modifying the DAVIS ROS driver. Because the CPU power of Xavier NX is limited, we decrease the image processing rate from 30 (for dataset test) to 7.5 in VINS-Mono, which decreases the performance a lot. The experiment result shows that the mean position error is 11.9%, and the mean yaw error is 2.3 deg/m. Figure 11 shows the position drift of the proposed method when the UAV flying.

---

[6] https://github.com/PX4/PX4-Autopilot

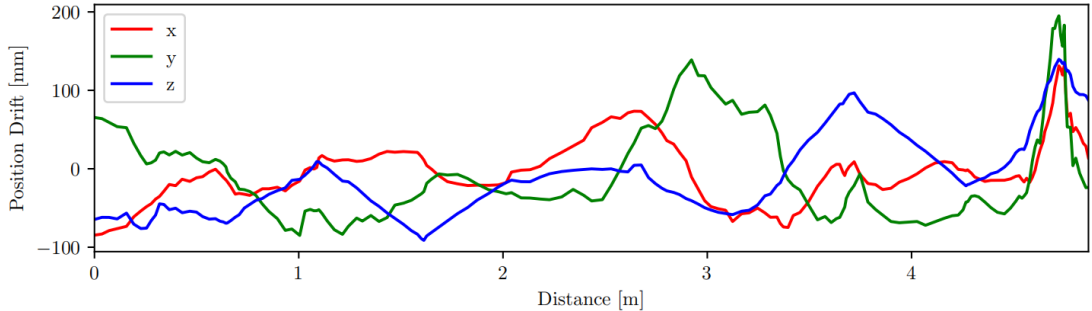

Figure 11: Position drift of the proposed method when the UAV flying

Event-based SLAM is mainly used when high speed motion, so the high processing rate of the SLAM system is important. VINS-Mono is an optimization-based SLAM, which needs relatively large computing power to realize better performance. When the computing power is limited, filter-based SLAM method can be better for event-based SLAM.

## 6 CONCLUSION

A method to adjust event accumulator settings for event-based SLAM is proposed. We analyze slice method, processing method for no motion, polarity usage, decay function and event contribution according to the event generation model and the property of the SLAM system, and propose the method to adjust them. For experiment verification, accumulated event frames are fed to VINS-Mono to construct an event-based SLAM system. A comparison between our approach and the state-of-the-art event frame based SLAM algorithm is conducted on public dataset. The result shows that our approach is better than the current state-of-the-art algorithm on 8 sequences and matched on 2 sequences. We also conducted UAV flight experiments to test the performance on an embedded system. Future work will focus on the fusion of tradition frames and events, which can obtain higher quality frames for SLAM systems.

## ACKNOWLEDGMENTS

Thanks to Kehan Xue for supplying dv_ros SDK[7], the research team of Professor Xiangke Wang from National University of Defense Technology for providing the DAVIS346 event camera, and Beijing Linkstech Co., Ltd. for providing the UAV and the laboratory.

---

[7] https://github.com/kehanXue/dv_ros